\pdfoutput=1

\documentclass[11pt]{article}

\usepackage{ACL2023}

\usepackage{times}
\usepackage{latexsym}

\usepackage[T1]{fontenc}

\usepackage[utf8]{inputenc}

\usepackage{microtype}

\usepackage{inconsolata}

\usepackage{booktabs}
\usepackage{url}
\usepackage{times}
\usepackage{latexsym}
\usepackage{hyperref}

\usepackage{amsmath,amssymb}
\usepackage{graphicx}
\usepackage{tabularx}
\usepackage{multirow}
\usepackage{pifont}
\usepackage{soul}

\usepackage[toc,page]{appendix}
\usepackage{xcolor}
\usepackage{floatrow}
\usepackage{makecell}
\usepackage{xspace}
\usepackage{colortbl}

\newfloatcommand{capbtabbox}{table}[][\FBwidth]

\usepackage{arydshln}

\usepackage{minibox}
\usepackage{tabularx,ragged2e}
\usepackage{tabularx} %
\usepackage{algorithm}
\usepackage[noend]{algpseudocode}
\usepackage{bbm}

\DeclareMathOperator{\softmax}{softmax}
\DeclareMathOperator{\leafs}{leaf-nodes}

\newcommand\ti[1]{\textit{#1}}

\newcommand\tf[1]{\textbf{#1}}

\definecolor{aquamarine}{rgb}{0.5, 1.0, 0.83}

\newcommand{\CC}{\cellcolor{gray!15}}
\newcommand{\CCC}{\cellcolor{gray!30}}
\definecolor{HLGreen}{HTML}{C1F7BB}
\definecolor{HLRed}{HTML}{FFC3C3}

\providecommand{\hl}[1]{
    {\protect\color{purple}{[Huihan: #1]}}
}

\newcommand{\xtask}{t}
\newcommand{\xconst}{c}
\newcommand{\xdemos}{E}
\newcommand{\xdemo}{e}
\newcommand{\constword}{\bar{c}}
\newcommand{\topicword}{\bar{t}}
\newcommand{\xin}{\mathcal{I}}
\newcommand{\ccheck}{\mathcal{C}}
\newcommand{\tcheck}{\mathcal{T}}
\newcommand{\template}{\mathcal{G}}
\newcommand{\taskname}{\textsc{Cognac}}
\newcommand{\sysname}{\textsc{CognacGen}}

\newcommand{\binaryguide}{binary verifier}
\newcommand{\topkguide}{top-k token}
\newcommand{\textguide}{textual example}

\title{Controllable Text Generation with Language Constraints}

\author{Howard Chen \quad Huihan Li \quad Danqi Chen \quad Karthik Narasimhan\\
  \large{Department of Computer Science, Princeton University}\\
  \texttt{\{howardchen, danqic, karthikn\}@cs.princeton.edu}\\
  \texttt{huihanl@usc.edu}
}

\begin{document}
\maketitle

\begin{abstract}

We consider the task of text generation in language models with constraints specified in natural language. To this end, we first create a challenging benchmark {\taskname}\footnote{{\taskname} stands for \tf{Co}ntrollable \tf{g}eneratio\tf{n} with l\tf{a}nguage \tf{c}onstraints.} 
that provides as input to the model a \textit{topic} with example text, along with a \textit{constraint} on text to be avoided. Unlike prior work, our benchmark contains knowledge-intensive constraints sourced from databases like Wordnet and Wikidata, which allows for straightforward evaluation while striking a balance between broad attribute-level and narrow lexical-level controls. We find that even state-of-the-art language models like GPT-3 fail often on this task, and propose a solution to leverage a language model's own internal knowledge to guide generation. Our method, called \sysname{}, first queries the language model to generate guidance terms for a specified topic or constraint, and uses the guidance to modify the model's token generation probabilities. We propose three 
forms of guidance (\binaryguide{}, \topkguide{}, \textguide{}), and employ prefix-tuning approaches to distill the guidance to tackle diverse natural language constraints.
Through extensive empirical evaluations, we demonstrate that \sysname{} can successfully generalize to unseen instructions and outperform competitive baselines in generating constraint conforming text.\footnote{Code and data are available at \url{https://github.com/princeton-nlp/Cognac}.}

\end{abstract}

\section{Introduction}\label{sec:intro}

As language models (LMs) become increasingly good at generating text indistinguishable from human writing, a key question emerges: `How can we best control them to produce what is required while preventing unwanted generations?' This is especially critical for reducing issues of toxicity and bias~\cite{gehman2020toxicprompt, xu2021minority, perez2022redteaming} and misinformation~\cite{taylor2022galactica} in applications that build on these models. %
Prior work has used special control codes~\cite{keskar2019ctrl} to steer the model towards generating text on certain topics, explored the use of classifiers at inference time to modify the LM's probability distribution~\cite{dathathri2020pplm,krause2021gedi,liu2021dexperts}, or prompting the LM itself to diagnosis and remove bias~\cite{schick2021diagnosisdebiasing}. While the former requires additional training with control codes, the other two approaches have only been shown to work with a small set of attributes as constraints.

\begin{figure}[t]
    \centering
    \includegraphics[width=1.0\textwidth]{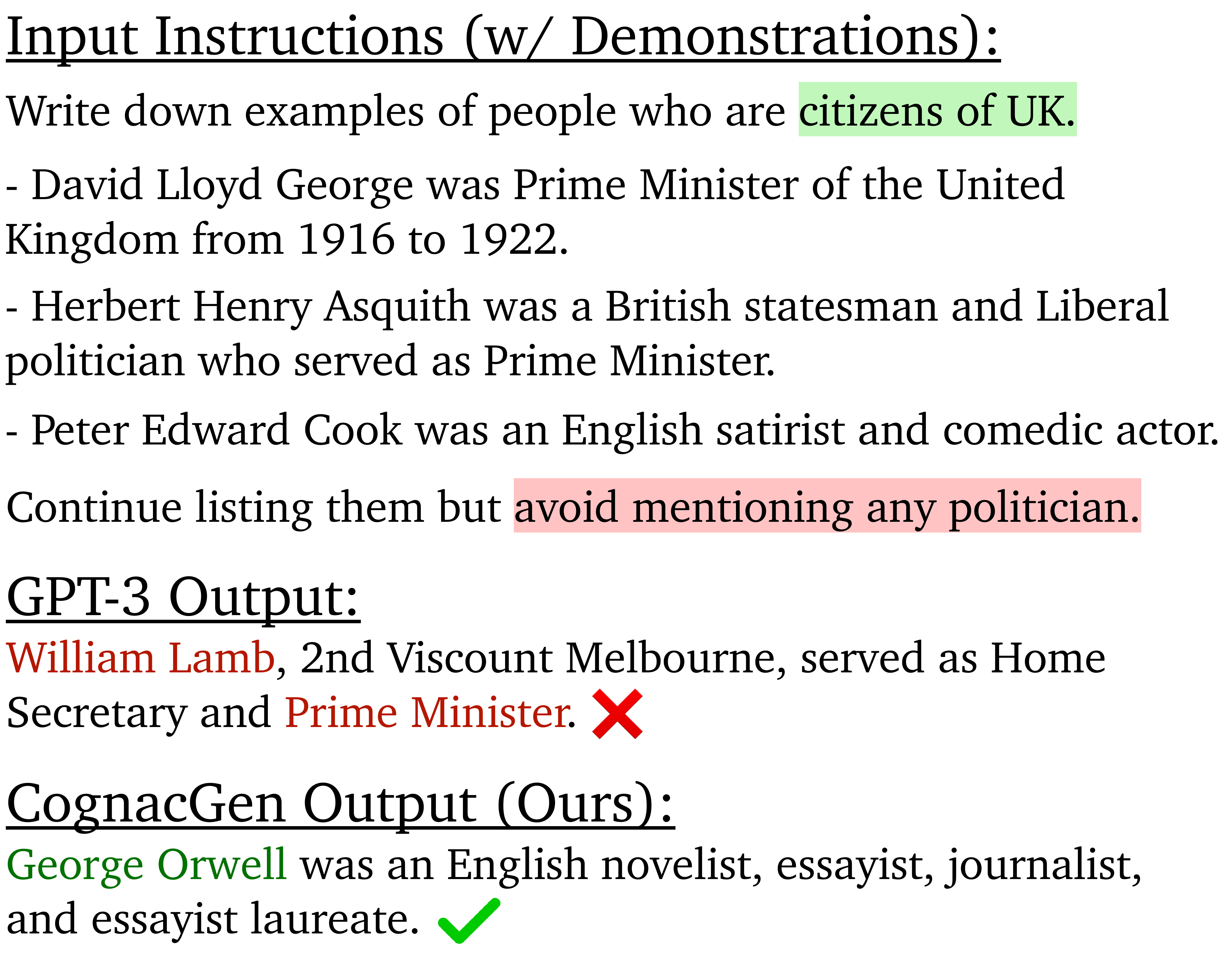}
    \caption{
        Constraining instructions and model generations.
        Green highlight specifies the topic to be covered. Red highlight specifies the constraint to conform to.
        GPT-3 generates continuation that mentioned a politician, thus violating the constraint.
        \sysname{} generates continuation that satisfies both the topic requirement and the constraint.
    }
    \label{fig:teaser}
\end{figure}

\begin{table*}[t]
    \centering
    \resizebox{0.95\columnwidth}{!}{%
    \begin{tabular}{cp{0.75\textwidth}p{0.25\textwidth}}
    \toprule
    \tf{Dataset} & \tf{Instruction w/ Demonstrations} & \tf{Topic / Constraint}\\
    \midrule
    \multirow{5}{*}{WordNet} & \sethlcolor{HLGreen}\hl{Talk about motor vehicle:} & \textbf{Topic}: motor vehicle\\
    & - A cruiser is a type of warship. & \textbf{Constraint}: car\\
    & - In motorsport, a safety car, or a pace car, is an automobile which [...] & \\
    & - A gas guzzler, in informal language, is a vehicle that is [...] & \\
    & \sethlcolor{HLRed}\hl{Do not talk about car}: & \\
    \midrule
    & - Arthur Neville Chamberlain [...] was a British politician [...] & \textbf{Topic}:\\
    \multirow{3}{*}{Wikidata} & - Maurice Harold Macmillan, [...] was a British Conservative statesman [...] & citizenship = UK\\ 
    & - Peter Edward Cook [...] was an English satirist and comedic actor [...] & \textbf{Constraint}:\\ 
    & \sethlcolor{HLGreen}\hl{The above are sentences describing people who are citizens of United Kingdom.} \sethlcolor{HLRed}\hl{Now write similar sentences as the above while omitting any mention of politician.} & occupation = politician\\
    \bottomrule
    \end{tabular}
    }
    \caption{Examples of the instruction task for WordNet and Wikidata. The instruction is specified by a topic ({\sethlcolor{HLGreen}\hl{green}}), a constraint ({\sethlcolor{HLRed}\hl{red}}), and a set of demonstration examples that are examples under the topic. The topic and constraint are specified by the corresponding entities (see details in \S\ref{sec:datasets}). Note that the position of the topic and the constraint with regard to the demonstrations may vary. 
    }
    \label{table:dataset_examples}
\end{table*}

In this work, we consider the problem of controlling generation in LMs with constraints specified in natural language (Figure~\ref{fig:teaser}). Our framework allows for the use of both guidance \emph{topics} that instructs the model on \emph{what to generate}, as well as \emph{constraints} that specifies \emph{what not to generate}, all described in plain English.\footnote{Although we focus on English, our techniques should generalize to other languages too.}  The use of natural language allows for better scalability (since new concepts can be expressed in English), ease of specification by end users of the model, and coverage of knowledge-intensive concepts, while not requiring any special retraining of the LM itself. We create a new benchmark called \taskname{} 
for this task containing two datasets based on WordNet~\cite{miller1995wordnet} and Wikidata~\cite{vrandecic2014wikidata}. These datasets contain knowledge-focused constraints that strike a balance between broad attribute-level and narrow lexical-level controls, while allowing for easy evaluation of constraint conformation. We find that even state-of-the-art LMs fail to follow simple language constraints. Figure~\ref{fig:teaser} shows an example of how GPT-3~\cite{brown2020gpt3} ignores the directive of not mentioning politicians (in red). 

To mitigate this failure, we develop \sysname{}, a language model generation method that can follow linguistic guidance and does not require any retraining of off-the-shelf LMs. \sysname{} uses prefix-tuning~\cite{li-liang-2021-prefix} over a copy of the same LM to distill from a guidance model that can generate both topic- and constraint-related words given natural language specifications, which can then be used at inference time to modify the output probabilities of the LM for controlled generation. We develop three types of guidance models---\binaryguide{}, \topkguide{} generator, and \textguide{} generator---that provide various levels of guidance to the LM. To handle the multi-token nature of the guidance examples, we also utilize a trie-based generation mechanism to track the guidance progress and ensure faithful guidance.

Our results show that \sysname{} outperforms prior methods and other strong baselines by a significant margin in our instruction conformance score metric, while keeping the generations fluent. 
When the topic and constraint are explicitly given (e.g., \ti{UK} and \ti{politican}; see Table~\ref{table:dataset_examples}), \sysname{} outperforms previous methods for controlled generation by up to $12$ points. Furthermore, \sysname{} leads $10$ points ahead of the prominent GPT-3 (\texttt{davinci}) model on both datasets when evaluating with natural language instructions. 
Our analysis shows that \sysname{} is able to improve generation even with imperfect guidance and can successfully generalize to unseen instructions.

\section{The {\taskname} Benchmark}\label{sec:setup}

\subsection{Task Setup}

We study the problem of conditional text generation with topics and constraints provided in natural language. As input, each context includes 1) a \textit{topic} to generate text on (e.g., ``List examples of people who are citizens of United Kingdom''), 2) a number of example generations under that topic (demonstrations) and 3) a \textit{constraint} that specifies what the model should not generate (e.g. ``Keep listing examples below, but do not mention any politician.'')---all specified in natural language. The goal is to train LMs to generate fluent on-topic content while respecting the specified constraint.

LMs typically learn a probability distribution $p_\theta(x)$ on sequences of tokens. An autoregressive LM can generate text by predicting the probability of the next token conditioned on the previous tokens: $p_\theta(x_j \mid x_{<j})$. 
In our task, we consider the previous tokens in the context to include a task specification $\xtask$, demonstrations $\xdemos = \{ \xdemo_k \}_{k=1}^K$, and a constraint $c$. We assume that the task description $\xtask$ is based on a topic entity $\topicword$. For example, ``Talk about sports'' is based on the topic entity ``sports''. 
Similarly, the constraint text $c$ is generated based on a constraint entity $\constword$.
The topic and constraint entities are added to the demonstrations using a template (\S\ref{sec:datasets}) into a full instruction $\xin = \template(\xtask, \xconst, \xdemos)$. \footnote{In our task, the demonstrations $\xdemos$  always share the same topic $\topicword$, yet they may violate the constraint $\constword$.}
This allows us to check the validity of each generation using a constraint checker $\ccheck (x, \constword) \in \{ 0, 1 \}$ and a topic checker $\tcheck(x, \topicword) \in \{ 0, 1 \}$. 
Specifically, a sequence $x$ generated by the LM is deemed valid when $x \sim p_\theta(x \mid \xin)$ such that $\ccheck(x, c) = 1$ (constraint conformed) and $\tcheck(x, t) = 1$ (on topic).
We show in Table~\ref{table:dataset_examples} examples of instructions and their corresponding topic and constraint entities.

Our task is challenging for three key reasons: 1) the model has to understand the topic and constraint specified in natural language, and 2) the topics and constraints are knowledge-intensive---broader than lexical-level constraint (e.g., `Include words ``car'' and ``drive'' in a sentence.')
yet more specific than broad attributes such as toxicity or sentiment, and 3) it has to respect both the topic (which specifies what to generate) and the constraint (which specifies what not to generate) simultaneously.

\begin{figure*}[t]
    \centering
    \includegraphics[width=0.97\textwidth]{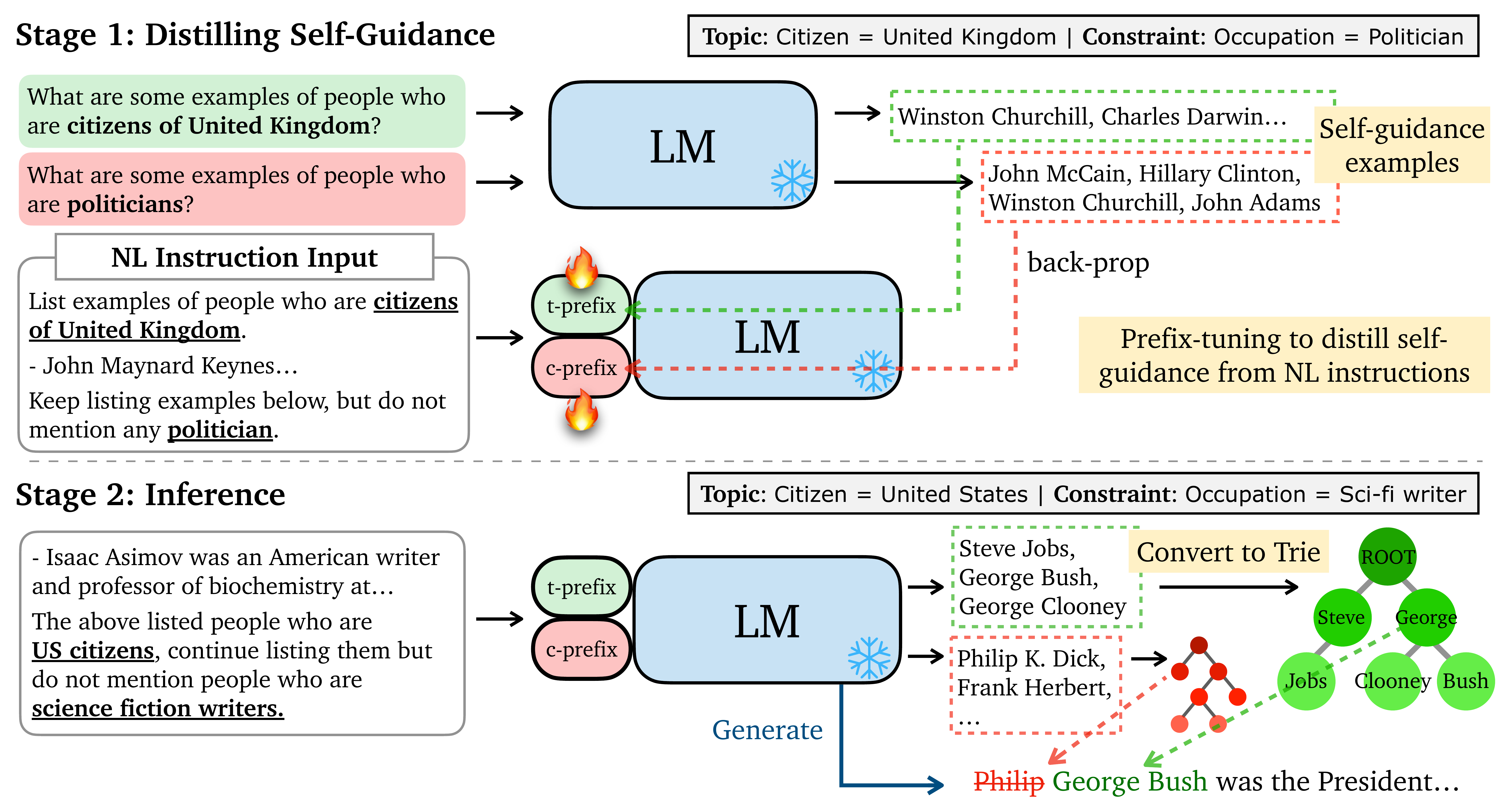}
    \caption{
        The two stages of \sysname{} with \textguide{} as guidance.
        Stage 1: the LM generates a list of guidance examples from the queries that specify the topic and constraint. During self-guidance distillation, the topic and constraint prefixes are tuned using the guidance example as target and the instruction with demonstrations as input. 
        Stage 2: The guidance model (blue LM \& the tuned prefixes) generates guidance examples from the test instance. The guidance examples are used to construct trie trees for both the topic (green) and the constraint (red).
        The generation (blue) LM's next token probability is modified by the tries. 
    }
    \label{fig:method} 
\end{figure*}

\subsection{Dataset Collection}\label{sec:datasets}

To our knowledge, there do not exist datasets for our task that contain topic and constraint specifications in natural language. Therefore, we create two new datasets based on WordNet and Wikidata for our {\taskname} benchmark. 

\paragraph{WordNet.}
We use WordNet \cite{miller1995wordnet} and its hypernymy relation to construct a hierarchical constraint dataset. We select five root nodes ``animal'', ``vehicle'', ``art'', ``food'', and ``sport'' from which the hierarchical structure is constructed. The leaf nodes are instances of their ancestors and are used as the topic and the constraint checker.
Concretely, when evaluating the generated text $x$ against a constraint entity $\constword$ using the WordNet constraint checker: $\ccheck^{\text{wordnet}} (x, \constword) =$ $\mathbbm{1}[\exists s \in \leafs(\constword): \mathcal{M}(s, x) = 1]$, where $\mathcal{M}(\cdot)$ denotes whether $s$ is a substring of $x$.\footnote{We implement the checks using exact match including multi-token entities.}
We sample two nodes as the topic and the constraint entities within the same subtree ($\texttt{higher-level}: \text{``vehicle''}, \texttt{lower-level}: \text{``car''}$ ) from the WordNet hierarchy, where the higher-level node is the topic and the lower-level node is the constraint. 
We collect a total of $221$ unique topics, $1,440$ unique constraints, and they form $3,073$ unique topic-constraint pairs.
We sample three leaf nodes under the topic node and use them as demonstrations ($\lvert \xdemos \rvert = 3$), where the demonstration is the first sentence from its Wikipedia page.
We collect a dataset of train/develop/test split of $3,000/ 500 / 500$. 

\paragraph{Wikidata.}
We also use Wikidata~\cite{vrandecic2014wikidata} to construct a second dataset. Each property and value pair (e.g., $\texttt{property}: \text{``citizenship''}$, $\texttt{value}: \text{``United Kingdom''}$; shown in Table~\ref{table:dataset_examples}) contains a set of names (e.g., Winston Churchill).
We use $5$ properties: \texttt{occupation}, \texttt{citizenship}, \texttt{education}, \texttt{birthplace}, and \texttt{deathplace} from Wikidata.
In each instance, the topic entity is a sampled property-value pair and the demonstraitons $\lvert \xdemos \rvert = 3$ are from the property-value name set. The constraint entity is selected by choosing what the GPT2-XL \cite{radford2019gpt2} most likely to generate.
When evaluating a generation $x$ with constarint entity $\constword$, the Wikidata constraint checker $\ccheck^{\text{wikidata}}(x, \constword) = \mathbbm{1}[\exists s \in \text{name-set}(\constword): \mathcal{M}(s, x) = 1]$.
We scrape from Wikipedia the corresponding first sentence of each entity.
We collect a total of $150$ unique topics, $261$ unique constraints, and they form $540$ unique topic-constraint pairs.
We collect a dataset of train/develop/test split of $1500 / 500 / 198$ examples.

We provide detailed data generation procedure for WordNet and Wikidata in \ref{sec:datagen}. Using the WordNet and Wikidata databases for the checker functions enjoys the benefit of straightforward and automatic evaluation. However, we recognize the knowledge bases come with their fundamental limits to capture all relevant entities.

\paragraph{Diverse natural language instructions.}
Our goal is to assess the model's ability to understand instructions that are diversely verbalized. For example, templates include instructions where the order of the topic and constraint vary and the lexical context differs.
We collect $35$ unique templates.
to reflect the diverse nature of the instructions and generate a total of $107,555$ and $18,900$ unique instructions for WordNet and Wikidata, respectively. We split them across train/develop/test as $3/3/29$ templates. 
The templates are collected by PhD students writing the first nine seed templates, which are then expanded by paraphrasing using GPT-3 \cite{brown2020gpt3}. 
The paraphrased templates were edited through human inspection to ensure quality.
We provide examples in \S\ref{sec:templates}.

\subsection{Evaluation Metrics} \label{sec:eval_metrics}
To evaluate different generation methods of LMs, we use metrics that test for correctness and fluency of the generations. Correctness is measured by the model's ability to generate text that conforms to the constraint while staying on topic. Fluency is measured by the model's ability to generate text that is coherent and not overly repetitive or simply copying from the input.

\paragraph{Instruction Conformance (IC).}
The main metric we use is whether the generation $x$ conforms to the constraint $c$ while staying on topic $t$:
\begin{align*}
    \text{IC} = \sum_{(\topicword, \constword, \xdemos) \in \mathcal{D}} \frac{\mathbbm{1} [ \tcheck(x, \topicword) = 1  \cap~\ccheck(x, \constword) = 1 ]}{\lvert \mathcal{D} \rvert},
\end{align*}
where $\mathcal{D}$ is the evaluation dataset.
A higher IC score indicates that the model can generate text that conforms to the constraint while staying on topic.
We also report the on-topic score $\sum_{x \in \mathcal{D}} \frac{\mathbbm{1} [ \tcheck(x, \topicword) = 1 ] }{\lvert \mathcal{D} \rvert}$ (higher is better) and the constraint violation score $\sum_{x \in \mathcal{D}} \frac{\mathbbm{1} [ \ccheck(x, \constword) = 1 ] }{\lvert \mathcal{D} \rvert}$ (lower is better).

\paragraph{Copy-BLEU.}
We report the extent to which the generation undesirably copies from the demonstrations. The Copy-BLEU score is calculated by taking the maximum BLEU score between the generated text and the $\lvert \xdemos \rvert$ demonstrations. The lower the Copy-BLEU, the less the generation copies from the demonstrations, hence more desirable.

\paragraph{Repetition (Rep-n).}
We report the ratio of the $n$-gram repetition (lower is better) in the generated text (\textbf{Rep-n}) proposed in \citet {welleck2020unlikelihood}.

\paragraph{Perplexity (PPL).}
The perplexity of the generated text is calculated with respect to a pre-trained GPT2-XL model \cite{radford2019gpt2} on the generated sentence (lower is better).

\section{Method}\label{sec:method}

\subsection{Overview}

We posit that the due to the knowledge-intensive nature of \taskname{}, the model will benefit from an explicit use of its own knowledge by querying itself.
To this end, we explicitly factorize the conditional probability as opposed to leaving the onus of inference to the LM. 
The desired distribution:
\begin{align*}
\begin{split}
p(x \mid \xdemos, \xtask, \xconst) &\propto p(x \mid \xdemos) p(t, c \mid x, \xdemos) \\
&= p(x \mid \xdemos) p(\xtask \mid x) p(\xconst \mid x)
\end{split}
\end{align*}
can be modeled by three components: 1) $p(x \mid \xdemos)$, which is the probability conditioned only on the demonstrations $\xdemos$ and 2) $p(\xtask \mid x)$ that evaluates if the task is performed, and 3) $p(\xconst \mid x)$ that evaluates if the constraint is conformed.
The former is the \textit{generation model}, which be modeled with the original pre-trained LM reasonably well, as recent work demonstrates LMs' ability to perform in-context learning with task specification and in-context demonstrations.
We use the latter as a \textit{guidance model} to steer generation explicitly.

\subsection{Guided Generation}\label{sec:guided_generation}
\sysname{} updates the next token prediction probability from the generation model by modifying the logits using the guidance (the ``Generate'' step in Figure~\ref{fig:method}). Specifically, the next token probability is modified as 
\begin{align*}
p(x_j \mid x_{<j}, \xin) = \softmax(o_j + \alpha o_j^\xtask - \beta o_j^\xconst ), \label{eq:gen_prob}
\end{align*}
where $s_j$ is the logits corresponding to the original probability $p(x_j \mid x_{<j}, E)$, $o_j^\xtask, o_j^\xconst \in \{0, 1\}^{\lvert V \rvert}$ are the \textit{guidance logits} provided by the guidance model at each generation step $j$, and $\alpha$ and $\beta$ are the hyperparameters that control the strength of the guidance.
We use greedy decoding to generate from the above probability for \sysname{} in our experiments. We describe how guidance logits are obtained in the following sections.

\subsection{Guidance Model}\label{sec:guidance_model}
Given a topic $\xtask$ or a constraint $\xconst$, we construct a guidance model that modify the guidance probabilities $p(\xtask \mid x)$ or $p(\xconst \mid x)$. The guidance model has the same architecture as the generation language model. We use the guidance model to produce a \textit{guidance logits} 
that modifies the next token logits of the language model at subword token indicies described in \S\ref{sec:guided_generation}. 

We explore three variations of guidance model:
1) \textit{\binaryguide{}}, 2) \textit{\topkguide{}}, and 3) \textit{\textguide{}}. 
All guidance models compute the guidance probability $p_\text{guide}(\cdot \mid q)$, where $q$ is a query based on a predefined template. The query template takes the constraint entity $\constword$ as input. We use $\constword = \text{`wine'}$ as an example throughout.

\begin{table}[t]
    \centering
    \resizebox{1.0\columnwidth}{!}{%
    \begin{tabular}{lll}
    \toprule
     & Query $q$ & Guidance Logit Tokens \\
    \midrule
    Binary & Is [pinot noir] & \texttt{If} $P(\text{`yes'} \mid q) > P(\text{`no'} \mid q)$, \\
    Verifier & a type of [wine]? & \texttt{then} \{pinot, noir\}\\
    \midrule
    Top-K & What are some & \texttt{top-k}\big(~$p_\text{guide}(\cdot \mid q)$~\big)\\
    Token & examples of [wine]? & = \{merlot, malbec\} \\
    \midrule
    &  & Initialize \texttt{Trie} with: \{merlot,\\
    Texual & What are some & cabernet, pinot noir, pinot gris\} \\
    Example & examples of [wine]? & \texttt{Trie}(\text{root}) \\
    & & $=$ \{merlot, cabernet, pinot\}\\
    & & \texttt{Trie}(\text{pinot}) =  \{noir, gris\}\\
    \bottomrule
    \end{tabular}}
    \caption{Guidance types, the corresponding query, and the final tokens used to construct the guidance logits. 
    }
    \label{table:guide_queries}
\end{table}

\paragraph{Binary verifier.} %
The \binaryguide{} evaluates the probability $p_{\text{guide}}(\text{``yes''} \mid q)$, where $q = \text{``Is}~x_j~\text{a type of}~\constword~\text{?''}$, where $x_j$ is the token to generate at timestep $j$.
Since sometimes $x_j$ does not carry clear meaning as a single token, we first perform a greedy decoded look-ahead \cite{lu2021neurologic} using the generation LM to construct a multi-token $x_{j:j+M}$ and obtain guidance from the verifier model \footnote{We use SpaCy part-of-speech parser to detect noun, noun phrase, or names. Therefore, $M$ is determined by the parser.}. 
Therefore, instead of using ``noir'' as the query word, we set $w = \text{``pinot noir''}$ to construct the query and send to the verifier model.
When $p_\text{guide}(\text{``yes''} \mid q) > p_\text{guide}(\text{``no''} \mid q)$, the generated entity $w$ is tokenized to construct a verifier guidance logits $o_j^c$, where its $i$-th index is $\mathbbm{1}[i \in \{ x_j, x_{j+1}, \dots, x_{j+M} \}]$.

The binary classifier guidance model can be viewed as an approximation of the constraint checker $\ccheck(\cdot, \cdot)$ but rely only on the existing knowledge in the LM.

\paragraph{Top-k token.}
The \topkguide{} guidance uses the next token probability distribution from the guidance model $p_{\text{guide}}(\cdot \mid q)$, where $q = \text{``What are some examples of wine?''}$.
Concretely, we use the top-$k$ tokens of this probability as guidance and construct the top-$k$ guidance logits as $o_j^c$, where its $i$-th index is $\mathbbm{1}[i \in \texttt{top-k}(p_{\text{guide}}(\cdot \mid q)) ]$. This variant falls short on providing guidance for multi-token entities due to its single-step nature (more discussion in \S\ref{sec:method_detail}).

\paragraph{Textual example.}
The \textguide{} guidance model takes a query $q = \text{``What are some examples of wine?''}$ and generates a set of guidance examples 
such as ``cabernet'', ``merlot'', ``pinot noir'', ``pinot gris''. We use top-$p$ \cite{holtzman2019curious} sampling with beam search to generate a diverse set of guidance examples. 
Directly tokenizing the examples into a set of subword tokens and use them to modify the logits might lead to suboptimal generation due to loss of order. For instance, a guidance example ``pinot noir'' may be split into ``pinot'' and ``noir'', and the probability of the two tokens will be modified in the same timestep. 

To mitigate the issue, we propose a trie tree based approach to decide what guidance to apply at each step. We construct a trie tree $\Gamma$ based on the generated guidance examples. With the above guidance examples, the root node will connect to its children nodes ``cabernet'', ``merlot'', and ``pinot''. The node ``pinot'' will connect to its children nodes ``noir'' and ``gris''.
At each generation step $j$, the trie tree takes the last generated token and return \textit{only} its children node tokens as guidance. For instance, if ``pinot'' is the current token to generate, the returned set of tokens are ``noir'' and ``gris'' as guidance.
We show the same procedure in Figure~\ref{fig:method} with names as an example.
This set of tokens is used to construct the textual guidance logits $o_j^c$, where its $i$-th index is $\mathbbm{1}[i \in \Gamma(x_{j-1})]$.

We summarize the different guidance models and their corresponding elements in Table~\ref{table:guide_queries} and provide a more detailed description for the textual example guidance in Algorithm~\ref{algo:gen_and_guide} in \ref{appx:appendix}.

\subsection{Tackling Diverse Natural Language Instructions}
The method described so far assumes that the topic and constraint are given and can be used in a query template to obtain guidance. However, the full \taskname{} task requires reading the entire instruction and demonstrations as input. 
We propose to train the model to take natural language instruction and demonstrations and generate the guidance directly.
With the set of diverse instructions (described in \S\ref{sec:datasets}), the model needs to infer the topic and the constraint entities from the full input containing the instruction and demonstrations. We fine-tune the generation model using prefix-tuning \cite{li-liang-2021-prefix} on the model's \textguide{} generated examples. This can be thought of as distilling the model's own knowledge by mimicking the textual guidance as the output and generalizing the implicit topic and constraint inference to unseen instructions.
Formally, we fine-tune the added prefix weights and save the prefix activations of the fixed guidance model $p_{\text{guide}}(y \mid [\xin; P]; \theta, \phi)$, where $y$ are the examples generated by the \textguide{} model\footnote{We only show results for \textguide{} since it works best in our experiments, but the distillation procedure can be applied to all three guidance models.} 
, $\phi$ is the added fine-tuning weights to generate the activations ($\theta$ remains fixed throughout), and $P$ is the added prefix tokens. The fine-tuning objective minimizes the loss $\mathcal{L}(\phi) = - \sum_t \log p_{\text{guide}}(y_t^* \mid [\xin; P]; \theta, \phi)$. At the end of the training, only prefix activations $\phi(P)$ are saved.
This step distills the model's own knowledge and generalizes the implicit topic and constraint inference to unseen natural language instructions (Figure~\ref{fig:method} Stage 1).

\citet{schick2021diagnosisdebiasing} share a similar high-level idea to use the same model's ability to identify bias and modify its generation. The authors propose to self-debias by prompting the model to obtain a biased probability, and subtract the probability from the original generation probability.
However, our method focuses on a more knowledge-intensive task, which requires the guidance to provide specific knowledge instead of a broader detection of biases. Our task also requires staying on topic and avoid constraints \textit{at the same time}. This warrants a different design for $p(t \mid x)$ and $p(c \mid x)$, which leads to developing the three guidance models and their tailored decoding design (e.g., incorporating trie). Finally, our setting expands to inferring topic and constraint (not given as control codes or attributes) from natural language instructions.

\begin{table*}[t]
    \centering
    \resizebox{1.0\columnwidth}{!}{
        \begin{tabular}{lcccccc}
        \toprule
        \multicolumn{7}{c}{Evaluation with Control Codes} \\
        \midrule
        & \multicolumn{3}{c}{WordNet} & \multicolumn{3}{c}{Wikidata}\\
        & \multicolumn{2}{c}{\CC Correctness} & \CCC Fluency & \multicolumn{2}{c}{\CC Correctness} & \CCC Fluency\\
        \multirow{2}{*}{Model} & \multirow{2}{*}{IC $\uparrow$} & On-Topic $\uparrow$ / & Rep-1 / Rep-2 / & \multirow{2}{*}{IC $\uparrow$} & On-Topic $\uparrow$ / & Rep-1 / Rep-2 / \\
        &  & Violation $\downarrow$ & Copy-BLEU / PPL $\downarrow$ & & Violation $\downarrow$ & Copy-BLEU / PPL $\downarrow$ \\
        \midrule
        Fine-Tuning     & 10.2  & 77.6 / 67.4 & 0.19 / 0.04 / 0.10 / 58.9 & 9.6  & 29.8 / 34.3 & 0.21 / 0.09 / 0.06 / 42.4 \\
        Self-Debiasing  & 24.2 & 50.8 / 26.6 & 0.27 / 0.14 / 0.01 / 51.4 & 19.8 & 35.6 / 29.6 & 0.27 / 0.15 / 0.05 / ~~8.0 \\
        \multicolumn{7}{l}{\sysname{}} \\
        ~~- \binaryguide{}      & 28.0 & 67.6 / 39.8 & 0.30 / 0.11 / 0.02 / 65.2 & 22.8 & 34.2 / 23.6 & 0.29 / 0.15 / 0.04 / 12.1 \\
        ~~- \topkguide{}       & 36.0 & 53.8 / 17.8 & 0.28 / 0.11 / 0.06 / 53.4 & 25.4 & 41.2 / 27.0 & 0.25 / 0.11 / 0.06 / ~~9.8 \\ %
        ~~- \textguide{}    & \bf{36.2} & 61.8 / 25.6 & 0.30 / 0.12 / 0.06 / 47.9 & \bf{35.8} & 43.8 / 14.2 & 0.17 / 0.06 / 0.05 / ~~6.4 \\
        \bottomrule
        \end{tabular}
    }
    \caption{
        Evaluation results on the control code setting on the development set of WordNet and Wikidata.
        We report Correctness and Fluency metrics for both datasets and IC is the Instruction Conformance score. The Fine-Tuning baseline use CTRL-style \cite{keskar2019ctrl} training and Self-Debiasing is adapted from \citet{schick2021diagnosisdebiasing}. 
    }
    \label{table:main_results}
\end{table*}
\begin{table*}[t]
    \centering
    \resizebox{1.0\columnwidth}{!}{
        \begin{tabular}{lccccccc}
        \toprule
        \multicolumn{8}{c}{Evaluation with  Natural Language Instructions} \\
        \midrule
        & & \multicolumn{3}{c}{WordNet} & \multicolumn{3}{c}{Wikidata}\\
        & & \multicolumn{2}{c}{\CC Correctness} & \CCC Fluency & \multicolumn{2}{c}{\CC Correctness} & \CCC Fluency\\
        \multirow{2}{*}{Model} & \multirow{2}{*}{Size} & \multirow{2}{*}{IC $\uparrow$} & On-Topic $\uparrow$ / & Rep-1 / Rep-2 / & \multirow{2}{*}{IC $\uparrow$} & On-Topic $\uparrow$ /  & Rep-1 / Rep-2 / \\
        & & & Violation $\downarrow$ & Copy-BLEU / PPL $\downarrow$ & & Violation $\downarrow$ & Copy-BLEU / PPL $\downarrow$ \\
        \midrule
        GPT-2 XL        & 1.5B & 12.0 & 86.8 / 74.8 & 0.18 / 0.03 / 0.10 / 57.7 & 18.4 & 38.6 / 37.0 & 0.15 / 0.04 / 0.26 / 33.8 \\
        GPT-3 (\texttt{davinci}) & 175B & 22.4 & 57.0 / 34.8 & 0.20 / 0.04 / 0.01 / 39.9 & 20.2 & 25.2 / 11.1 & 0.08 / 0.01 / 0.01 / 22.2 \\
        \sysname{} & 1.5B & \bf{32.4} & 54.8 / 22.4 & 0.29 / 0.13 / 0.02 / 51.7 & \bf{31.8} & 43.9 / 19.7 & 0.22 / 0.10 / 0.02 / ~~9.6 \\
        \midrule
        InstructGPT & 175B & \bf{49.0} & 82.6 / 33.6 & 0.20 / 0.05 / 0.02 / 28.3 & \bf{41.9} & 52.5 / 16.7 & 0.07 / 0.01 / 0.02 / 15.7 \\
        \bottomrule
        \end{tabular}
    }
    \caption{
        Results on the NL instruction setting on the test set of WordNet and Wikidata. We report Correctness and Fluency metrics for both datasets and IC is the Instruction Conformance score.
        The natural language instruction templates do not overlap across train/development/test splits. \sysname{} uses \textguide{} guidance.
    }
    \label{table:main_results_2}
\end{table*}

\section{Experimental Setup}\label{sec:exp}

\paragraph{Evaluation}
We perform evaluations under two settings for both datasets in \taskname{}: 
\begin{enumerate}
    \item Both the topic and the constraint are specified using a \textbf{control code} each;
    \item The topic and the constraint are specified in the form of a \textbf{natural language instruction}.
\end{enumerate}
The control code setting allows us to better compare with prior work, which mostly uses a small set of attributes to steer generation. In this setting, we examine \sysname{} with all three guidance types: \binaryguide{}, \topkguide{}, and \textguide{}. We adapt \sysname{} to this setting by skipping the self-guidance distillation step and use the topic and constraint directly as control code.

However, the NL instruction setting is more realistic and closer to the real-world use case, where a user can control the LM with natural language.
For this setting, the test set split (\S\ref{sec:setup}) contains a set of unseen instruction templates that are never seen in the train set (details in \S\ref{sec:datasets}).
We use \textguide{} guidance for \sysname{} in this setting because we observe its superior performance across the board in the control code setting.

\paragraph{Baselines}
When evaluating with control codes, we compare \sysname{} to a fine-tuned model baseline built on CTRL \cite{keskar2019ctrl}, where the topic and the constraint are provided as control codes that are appended at the beginning of the input text. 
We also compare to the self-debiasing technique proposed in \cite{schick2021diagnosisdebiasing}, as it is the only method in the recent controllable generation approach that can apply to arbitraty number of control codes/attributes without fine-tuning.
To adapt \sysname{} to the control code setting, we can simply skip the self-guidance distillation stage and use the topic and constraint as control.

When evaluating with natural language instructions, we compare with two large language models (175B parameters): GPT-3 (\texttt{davinci}) \cite{brown2020gpt3}, and InstructGPT (\texttt{text-davinci-002}) \cite{ouyang2022instructgpt}.

\paragraph{Model details}
All of our \sysname{} variants use GPT-2 XL (1.5B parameters) \cite{radford2019gpt2} for both generation and guidance models.
The \topkguide{} uses top $20$ tokens for topic and top $40$ tokens for the constraint. The \textguide{} guidance generates $200$ tokens for building the trie.
For both GPT-3 and InstructGPT, we use top-$p = 0.95$ and temperature $\tau = 0.9$.
We provide more details about self-guidance distillation training details in \S\ref{sec:method_detail}.

\section{Results}\label{sec:results}

\paragraph{Main results.}
Tables~\ref{table:main_results} and \ref{table:main_results_2} display the results for the two evaluation settings, respectively. In the control code setting (Table~\ref{table:main_results}), \sysname{} (\textguide{}) achieves the best instruction conformance (IC) scores, outperforming the self-debiasing baseline by $12$ points on WordNet and by $16$ points on Wikidata. The fine-tuned baseline achieves the lowest IC scores across both datasets. Among \sysname{}'s variants, \textguide{} guidance performs better than \topkguide{} guidance and the \binaryguide{}. All model variants of \sysname{} seem to be equally fluent, with \sysname{} \textguide{} having a desirable slightly lower $47.9$ perplexity. 

In the NL instruction setting (Table~\ref{table:main_results_2}), \sysname{} \textguide{} achieves a higher performance than GPT-3 (legacy) by $10.0$ points on WordNet and $11.6$ points on Wikidata, despite having much fewer parameters (1.5B vs 175B).
InstructGPT achieves much higher scores (49 IC on Wordnet and 41.9 IC on Wikidata), but it is also a much larger model(175B)  and is also fine-tuned on instruction following using human feedback (RLHF)~\cite{ouyang2022instructgpt}.

To analyse model performance on different kinds of templates, we report IC scores for each of the three templates in development sets in Table~\ref{table:nl-instruct-generalize}.
We observe that performance among different templates stays about the same for WordNet, but for Wikidata, the template with the topic and constraint specified at the end
proves to be more challenging than others. This highlights challenges due to structural variations in instruction templates and how this may manifest differently in each dataset.

\begin{table}[t]
    \centering
    \resizebox{0.95\columnwidth}{!}{%
    \begin{tabular}{lcc}
    \toprule
    Template Position & WordNet & Wikidata\\
    \midrule
    Beginning: topic; End: constraint & 33.2 & 32.6\\
    Beginning: topic \& constraint & 31.2 & 17.4\\
    End: topic \& constraint & 36.6 & 11.4\\
    \bottomrule
    \end{tabular}}
    \caption{
        Instruction Conformance for different natural language templates using \sysname{} \textguide{}. These three templates are applied to all the instances in the development set. 
    }
    \label{table:nl-instruct-generalize}
\end{table}
\vspace{-12pt}

\paragraph{Performance analysis by category.}
We analyze the performance of \sysname{} (\textguide{}) by category for both WordNet (Table~\ref{table:wordnet_breakdown}) and Wikidata (Table~\ref{table:wikidata_breakdown}), revealing how each category provides different challenges. 
We observe that \sysname{} struggles to avoid violating constraints for the `Art' category at a IC of $15.0$, a much lower score compared to other categories which all have $>30.0$ IC.
Moreover, for knowledge-heavy categories such as `Art' in WordNet and `birthplace'/`deathplace' (as topic) in Wikidata, \sysname{} struggles to stay on topic. 

\begin{table}[t]
    \centering
    \resizebox{0.85\columnwidth}{!}{%
        \begin{tabular}{lccc}
        \toprule
        & \multicolumn{3}{c}{WordNet} \\
        & IC $\uparrow$ & On-Topic $\uparrow$ & Violation $\downarrow$ \\
        \midrule
        Animal & 33.9 & 62.8 & 29.0\\
        Vehicle & 43.0 & 78.0 & 35.0\\
        Food & 36.0 & 64.6 & 28.7\\
        Sport & 36.4 & 66.7 & 30.3\\
        Art & 15.0 & 55.0 & 40.0 \\
        \bottomrule
        \end{tabular}
    }
    \caption{WordNet performance breakdown by category. In every example, the topic and constraint are coming from the same category.
    }
    \label{table:wordnet_breakdown}
\end{table}

\begin{table}[t]
    \centering
    \resizebox{0.95\columnwidth}{!}{%
        \begin{tabular}{lrrr}
        \toprule
        & \multicolumn{3}{c}{Wikidata} \\
        & IC $\uparrow$ & On-Topic $\uparrow$ & Vio. $\downarrow$ \\
        \midrule
        When Used as Topic & & & \\
        ~~- Occupation & 32.8 & 44.4 & 20.0 \\
        ~~- Citizen & 46.2 & 55.5 & 12.6 \\
        ~~- Education & 33.3 & 33.3 & 33.3 \\
        ~~- Birthplace & 0.0 & 0.0 & 18.2 \\
        ~~- Deathplace & 4.4 & 4.4 & 20.0 \\
        \midrule
        When Used as Constraint & & & \\
        ~~- Occupation & 15.8 & 29.0 & 23.7 \\
        ~~- Citizen & 14.1 & 26.3 & 37.4 \\
        ~~- Education & 44.4 & 44.4 & 5.6 \\
        ~~- Birthplace & 40.7 & 50.3 & 12.4 \\
        ~~- Deathplace & 25.5 & 29.1 & 3.6 \\
        \bottomrule
        \end{tabular}
    }
    \caption{Wikidata performance breakdown by category. In Wikidata, topic and constraint are often from different categories. On-topic only accounts for when the category is used as topic. Vio.: violation  only accounts for when the category is used as constraint.}
    \label{table:wikidata_breakdown}
\end{table}
\vspace{-5pt}

\paragraph{Model ablations.}
To provide more insight into the workings of \sysname{}, we ablate away the trie-based generation and also compare with a database oracle model on Wikidata, which provides an upper bound on IC score when using the proposed decoding method proposed in \S\ref{sec:guided_generation}. This oracle model has access to the knowledge base, and hence can provideperfect guidance (Table~\ref{table:trie_oracle_ablation}). 
The oracle achieves an IC of 73, compared to \sysname{}'s 35.8, indicating that there is quite a bit of room for improvement on our task, both in terms of generating more on-topic text and avoiding violations. Further, both \sysname{} and the oracle degrade in performance drastically when the tries are removed, highlighting the effectiveness of using tries to guide generation. This degradation is particularly pronounced due to the need for generating multi-token names in Wikidata.

\begin{table}[t]
    \centering
    \resizebox{1.0\columnwidth}{!}{%
        \begin{tabular}{lcccc}
        \toprule
        Model & IC $\uparrow$ & On-Topic $\uparrow$ & Vio. $\downarrow$ & PPL $\downarrow$\\
        \midrule
        \sysname{} & \bf{35.8} & \bf{43.8} & 14.2 & 6.4\\
        ~~~- w/o trie  & 10.4 & 11.2 & \bf{3.2} & \bf{5.0}\\
        \midrule
        Oracle     & \bf{73.0} & \bf{73.4} & \bf{0.4} & 9.9\\
        ~~~- w/o trie & 13.2 & 12.6 & 2.0 & \bf{3.8}\\
        \bottomrule
        \end{tabular}
    }
    \caption{Ablation on trie between \sysname{} (\textguide{}) and the oracle which assumes access to the knowledge base, instead of relying on the LM’s internal knowledge. The ablation is on Wikidata.
    }
    \label{table:trie_oracle_ablation}
\end{table}

\paragraph{Qualitative examples.}
Finally, Table~\ref{table:generation_examples} shows example generations from \sysname{} and GPT-3 (\texttt{davinci}) on WordNet and Wikidata. For WordNet, \sysname{} generates constraint comforming output yet GPT-3's generation violates the constraint by generating examples including scallop.
On Wikidata, \sysname{} is able to follow the instructions and generate a sentence about a journalist, while GPT-3 fails to stay on topic.

\section{Related Work}\label{sec:related}

\paragraph{Constrained text generation.}
Prior approaches to constrained text generation fall into several categories. First, works like CTRL~~\cite{keskar2019ctrl}, GeDi \cite{krause2021gedi} and Neurologic decoding~\cite{lu2021neurologic,lu2022neurologicAstar} use additional context information such as control codes, attributes or word lists to condition their generations. Second, papers like PPLM~\cite{dathathri2020pplm} and DExperts \cite{liu2021dexperts} modify the model's output probabilities during inference using classifiers and auxiliary models, respectively. Along the same lines, Unlikelihood training \cite{welleck2020unlikelihood} and CRINGE~\cite{adolphs2022cringe} use auxiliary token-level and sequence-level objectives to discourage models from assigning high probabilities to certain tokens or sequences, while  Quark~\cite{lu2022quark} and \citet{liu2021constrained} use reinforcement learning to do the same. All these approaches are limited by the type of control they exert over the language model, restricted to high-level concepts like sentiment, toxicity or repetition and usually employing a fixed set of pre-determined binary or categorical codes.

The third category consists of methods that use a language model's own knowledge to guide its generations, which is probably most similar to our work. This includes self-debiasing~\cite{schick2021diagnosisdebiasing}, which  reduces toxicity by prompting the model to generate toxic content and offset this behavior from the main generation LM. This method works is limited to a single high-level attribute (e.g. toxicity) that needs to be suppressed while \sysname{} can handle a composition of attributes (topic + constraints) based on precise factual knowledge. More recently, Self-correction~\cite{welleck2022selfcorrect} learns a correction module that iteratively edits generated text and is trained using scalar or language feedback.  Their method requires progressively training and updating the corrector module and the generation uses multiple iterations, whereas our guidance module is only prefix-tuned once and can generate text in one pass.

\paragraph{Instruction following.}
A large body of literature in embodied agent learning has focused on following instructions or constraints in a grounded setting~\citep{vogel2010learning,chen2011learning,artzi2013weakly,luketina2019survey,misra2018mapping,yang2020safe}.These papers focus on instruction understanding that maps to actions in an external environment, as opposed to text generation. More recently, papers have looked explored finetuning language models to follow instructions in natural language for various NLP tasks~\cite{ouyang2022instructgpt,mishra-etal-2022-cross,wang2022super,wei2021finetuned,bach2022promptsource}.
In contrast to our work, these methods do not focus on using language to control the generated text in a fine-grained manner and require costly fine-tuning or large-scale prompt creation.

\section{Conclusion}\label{sec:conclusion}
We have introduced a new task for controllable generation in language models with constraints specified in natural language. We developed \taskname{}, a new benchmark containing knowledge-based constraints using data from Wordnet and Wikidata and showed that even state-of-the-art language models like GPT-3 fail to conform to the provided instructions. We then develop \sysname{}, a method to use knowledge internal to a language model to guide its own generations. Our approach involves several key innovations such as guidance self-distillation using prefix-tuning and a trie-based decoding scheme based on the guidance of textual examples. This helps the model generate on-topic text that violates constraints less frequently compared to several baselines, including much larger models like GPT-3. More importantly, our method require training only the prefix parameters and can easily be scaled to larger models without requiring significant computational overhead. Our analysis also revealed that there is still significant room to improve on \taskname{} and we hope future approaches will find the benchmark useful for developing better methods to control language models.

\section*{Limitations}\label{sec:limitation}
Our work is aimed at reducing undesirable generations in LMs while promoting desirable text. A successful scenario would increase the instruction conformance score when our method is applied. However, our benchmark is limited by the comprehensiveness of the underlying knowledge bases (KB) used. Any generation that goes beyond the factual knowledge present in the KB would be deemed incorrect, which may amplify any bias existing in the KB, e.g., people with certain background or ethnicity might be underrepresented. Furthermore, even when the generation is within the scope of the KB, the model might still have a tendency to choose certain types of knowledge over another. These implicit biases might cause unfairness to the end users of the model.
\section*{Acknowledgements}
\label{sec:ack}
We thank the members of the Princeton NLP group for the helpful discussions. In particular, we thank Austin Wang and Tianyu Gao for their valuable feedback.

\bibliography{ref,instruction}

\begin{thebibliography}{34}
\expandafter\ifx\csname natexlab\endcsname\relax\def\natexlab#1{#1}\fi

\bibitem[{Adolphs et~al.(2022)Adolphs, Gao, Xu, Shuster, Sukhbaatar, and
  Weston}]{adolphs2022cringe}
Leonard Adolphs, Tianyu Gao, Jing Xu, Kurt Shuster, Sainbayar Sukhbaatar, and
  Jason Weston. 2022.
\newblock The cringe loss: Learning what language not to model.
\newblock In \emph{preprint}.

\bibitem[{Artzi and Zettlemoyer(2013)}]{artzi2013weakly}
Yoav Artzi and Luke Zettlemoyer. 2013.
\newblock \href {https://doi.org/10.1162/tacl_a_00209} {Weakly supervised
  learning of semantic parsers for mapping instructions to actions}.
\newblock \emph{Transactions of the Association for Computational Linguistics},
  1:49--62.

\bibitem[{Bach et~al.(2022)Bach, Sanh, Yong, Webson, Raffel, Nayak, Sharma,
  Kim, Bari, F{\'e}vry et~al.}]{bach2022promptsource}
Stephen Bach, Victor Sanh, Zheng~Xin Yong, Albert Webson, Colin Raffel, Nihal~V
  Nayak, Abheesht Sharma, Taewoon Kim, M~Saiful Bari, Thibault F{\'e}vry,
  et~al. 2022.
\newblock Promptsource: An integrated development environment and repository
  for natural language prompts.
\newblock In \emph{Proceedings of the 60th Annual Meeting of the Association
  for Computational Linguistics: System Demonstrations}, pages 93--104.

\bibitem[{Brown et~al.(2020)Brown, Mann, Ryder, Subbiah, Kaplan, Dhariwal,
  Neelakantan, Shyam, Sastry, Askell, Agarwal, Herbert-Voss, Krueger, Henighan,
  Child, Ramesh, Ziegler, Wu, Winter, Hesse, Chen, Sigler, Litwin, Gray, Chess,
  Clark, Berner, McCandlish, Radford, Sutskever, and Amodei}]{brown2020gpt3}
Tom~B. Brown, Benjamin Mann, Nick Ryder, Melanie Subbiah, Jared Kaplan,
  Prafulla Dhariwal, Arvind Neelakantan, Pranav Shyam, Girish Sastry, Amanda
  Askell, Sandhini Agarwal, Ariel Herbert-Voss, Gretchen Krueger, Tom Henighan,
  Rewon Child, Aditya Ramesh, Daniel~M. Ziegler, Jeffrey Wu, Clemens Winter,
  Christopher Hesse, Mark Chen, Eric Sigler, Mateusz Litwin, Scott Gray,
  Benjamin Chess, Jack Clark, Christopher Berner, Sam McCandlish, Alec Radford,
  Ilya Sutskever, and Dario Amodei. 2020.
\newblock Language models are few-shot learners.
\newblock In \emph{Advances in Neural Information Processing Systems
  (NeurIPS)}.

\bibitem[{Chen and Mooney(2011)}]{chen2011learning}
David~L. Chen and Raymond~J. Mooney. 2011.
\newblock \href {http://www.aaai.org/ocs/index.php/AAAI/AAAI11/paper/view/3701}
  {Learning to interpret natural language navigation instructions from
  observations}.
\newblock In \emph{Proceedings of the Twenty-Fifth {AAAI} Conference on
  Artificial Intelligence, {AAAI} 2011, San Francisco, California, USA, August
  7-11, 2011}. {AAAI} Press.

\bibitem[{Dathathri et~al.(2020)Dathathri, Madotto, Lan, Hung, Frank, Molino,
  Yosinski, and Liu}]{dathathri2020pplm}
Sumanth Dathathri, Andrea Madotto, Janice Lan, Jane Hung, Eric Frank, Piero
  Molino, Jason Yosinski, and Rosanne Liu. 2020.
\newblock Plug and play language models: A simple approach to controlled text
  generation.
\newblock In \emph{International Conference on Learning Representations
  (ICLR)}.

\bibitem[{Gehman et~al.(2020)Gehman, Gururangan, Sap, Choi, and
  Smith}]{gehman2020toxicprompt}
Samuel Gehman, Suchin Gururangan, Maarten Sap, Yejin Choi, and Noah~A. Smith.
  2020.
\newblock Realtoxicityprompts: Evaluating neural toxic degeneration in language
  models.
\newblock In \emph{Findings of the Empirical Methods in Natural Language
  Processing (EMNLP Findings)}.

\bibitem[{Holtzman et~al.(2019)Holtzman, Buys, Du, Forbes, and
  Choi}]{holtzman2019curious}
Ari Holtzman, Jan Buys, Li~Du, Maxwell Forbes, and Yejin Choi. 2019.
\newblock The curious case of neural text degeneration.
\newblock In \emph{International Conference on Learning Representations}.

\bibitem[{Keskar et~al.(2019)Keskar, McCann, Varshney, Xiong, and
  Socher}]{keskar2019ctrl}
Nitish~Shirish Keskar, Bryan McCann, Lav~R. Varshney, Caiming Xiong, and
  Richard Socher. 2019.
\newblock {CTRL}: A conditional transformer language model for controllable
  generation.
\newblock In \emph{preprint}.

\bibitem[{Krause et~al.(2021)Krause, Gotmare, McCann, Keskar, Joty, Socher, and
  Rajani}]{krause2021gedi}
Ben Krause, Akhilesh~Deepak Gotmare, Bryan McCann, Nitish~Shirish Keskar,
  Shafiq Joty, Richard Socher, and Nazneen~Fatema Rajani. 2021.
\newblock \href {https://aclanthology.org/2021.findings-emnlp.424} {{G}e{D}i:
  Generative discriminator guided sequence generation}.
\newblock In \emph{Findings of the Empirical Methods in Natural Language
  Processing (EMNLP Findings)}.

\bibitem[{Li and Liang(2021)}]{li-liang-2021-prefix}
Xiang~Lisa Li and Percy Liang. 2021.
\newblock \href {https://doi.org/10.18653/v1/2021.acl-long.353} {Prefix-tuning:
  Optimizing continuous prompts for generation}.
\newblock In \emph{Proceedings of the 59th Annual Meeting of the Association
  for Computational Linguistics and the 11th International Joint Conference on
  Natural Language Processing (Volume 1: Long Papers)}, pages 4582--4597,
  Online. Association for Computational Linguistics.

\bibitem[{Liu et~al.(2021{\natexlab{a}})Liu, Sap, Lu, Swayamdipta, Bhagavatula,
  Smith, and Choi}]{liu2021dexperts}
Alisa Liu, Maarten Sap, Ximing Lu, Swabha Swayamdipta, Chandra Bhagavatula,
  Noah~A. Smith, and Yejin Choi. 2021{\natexlab{a}}.
\newblock \href {https://aclanthology.org/2021.acl-long.522} {{DE}xperts:
  Decoding-time controlled text generation with experts and anti-experts}.
\newblock In \emph{Association for Computational Linguistics (ACL)}.

\bibitem[{Liu et~al.(2021{\natexlab{b}})Liu, Zhang, Han, Zhang, and
  Tu}]{liu2021constrained}
Yixian Liu, Liwen Zhang, Wenjuan Han, Yue Zhang, and Kewei Tu.
  2021{\natexlab{b}}.
\newblock Constrained text generation with global guidance--case study on
  commongen.

\bibitem[{Loshchilov and Hutter(2017)}]{loshchilov2017decoupled}
Ilya Loshchilov and Frank Hutter. 2017.
\newblock Decoupled weight decay regularization.
\newblock \emph{arXiv preprint arXiv:1711.05101}.

\bibitem[{Lu et~al.(2022{\natexlab{a}})Lu, Welleck, Jiang, Hessel, Qin, West,
  Ammanabrolu, and Choi}]{lu2022quark}
Ximing Lu, Sean Welleck, Liwei Jiang, Jack Hessel, Lianhui Qin, Peter West,
  Prithviraj Ammanabrolu, and Yejin Choi. 2022{\natexlab{a}}.
\newblock \href {https://arxiv.org/abs/2205.13636} {Quark: Controllable text
  generation with reinforced unlearning}.
\newblock In \emph{Advances in Neural Information Processing Systems
  (NeurIPS)}.

\bibitem[{Lu et~al.(2022{\natexlab{b}})Lu, Welleck, West, Jiang, Kasai,
  Khashabi, Bras, Qin, Yu, Zellers, Smith, and Choi}]{lu2022neurologicAstar}
Ximing Lu, Sean Welleck, Peter West, Liwei Jiang, Jungo Kasai, Daniel Khashabi,
  Ronan~Le Bras, Lianhui Qin, Youngjae Yu, Rowan Zellers, Noah~A. Smith, and
  Yejin Choi. 2022{\natexlab{b}}.
\newblock Neurologic a*esque decoding: Constrained text generation with
  lookahead heuristics.
\newblock In \emph{North American Association for Computational Linguistics
  (NAACL)}.

\bibitem[{Lu et~al.(2021)Lu, West, Zellers, Bras, Bhagavatula, and
  Choi}]{lu2021neurologic}
Ximing Lu, Peter West, Rowan Zellers, Ronan~Le Bras, Chandra Bhagavatula, and
  Yejin Choi. 2021.
\newblock Neurologic decoding: (un)supervised neural text generation with
  predicate logic constraints.
\newblock In \emph{North American Association for Computational Linguistics
  (NAACL)}.

\bibitem[{Luketina et~al.(2019)Luketina, Nardelli, Farquhar, Foerster, Andreas,
  Grefenstette, Whiteson, and Rockt{\"{a}}schel}]{luketina2019survey}
Jelena Luketina, Nantas Nardelli, Gregory Farquhar, Jakob~N. Foerster, Jacob
  Andreas, Edward Grefenstette, Shimon Whiteson, and Tim Rockt{\"{a}}schel.
  2019.
\newblock \href {https://doi.org/10.24963/ijcai.2019/880} {A survey of
  reinforcement learning informed by natural language}.
\newblock In \emph{Proceedings of the Twenty-Eighth International Joint
  Conference on Artificial Intelligence, {IJCAI} 2019, Macao, China, August
  10-16, 2019}, pages 6309--6317. ijcai.org.

\bibitem[{Miller(1995)}]{miller1995wordnet}
George~A Miller. 1995.
\newblock Wordnet: a lexical database for english.
\newblock \emph{Communications of the ACM}, 38(11):39--41.

\bibitem[{Mishra et~al.(2022)Mishra, Khashabi, Baral, and
  Hajishirzi}]{mishra-etal-2022-cross}
Swaroop Mishra, Daniel Khashabi, Chitta Baral, and Hannaneh Hajishirzi. 2022.
\newblock \href {https://doi.org/10.18653/v1/2022.acl-long.244} {Cross-task
  generalization via natural language crowdsourcing instructions}.
\newblock In \emph{Proceedings of the 60th Annual Meeting of the Association
  for Computational Linguistics (Volume 1: Long Papers)}, pages 3470--3487,
  Dublin, Ireland. Association for Computational Linguistics.

\bibitem[{Misra et~al.(2018)Misra, Bennett, Blukis, Niklasson, Shatkhin, and
  Artzi}]{misra2018mapping}
Dipendra Misra, Andrew Bennett, Valts Blukis, Eyvind Niklasson, Max Shatkhin,
  and Yoav Artzi. 2018.
\newblock \href {https://doi.org/10.18653/v1/D18-1287} {Mapping instructions to
  actions in 3{D} environments with visual goal prediction}.
\newblock In \emph{Proceedings of the 2018 Conference on Empirical Methods in
  Natural Language Processing}, pages 2667--2678, Brussels, Belgium.
  Association for Computational Linguistics.

\bibitem[{Ouyang et~al.(2022)Ouyang, Wu, Jiang, Almeida, Wainwright, Mishkin,
  Zhang, Agarwal, Slama, Ray, Schulman, Hilton, Kelton, Miller, Simens, Askell,
  Welinder, Christiano, Leike, and Lowe}]{ouyang2022instructgpt}
Long Ouyang, Jeff Wu, Xu~Jiang, Diogo Almeida, Carroll~L. Wainwright, Pamela
  Mishkin, Chong Zhang, Sandhini Agarwal, Katarina Slama, Alex Ray, John
  Schulman, Jacob Hilton, Fraser Kelton, Luke Miller, Maddie Simens, Amanda
  Askell, Peter Welinder, Paul Christiano, Jan Leike, and Ryan Lowe. 2022.
\newblock Training language models to follow instructions with human feedback.
\newblock In \emph{preprint}.

\bibitem[{Perez et~al.(2022)Perez, Huang, Song, Cai, Ring, Aslanides, Glaese,
  McAleese, and Irving}]{perez2022redteaming}
Ethan Perez, Saffron Huang, Francis Song, Trevor Cai, Roman Ring, John
  Aslanides, Amelia Glaese, Nat McAleese, and Geoffrey Irving. 2022.
\newblock Red teaming language models with language models.

\bibitem[{Radford et~al.(2019)Radford, Wu, Child, Luan, Amodei, and
  Sutskever}]{radford2019gpt2}
Alec Radford, Jeff Wu, Rewon Child, David Luan, Dario Amodei, and Ilya
  Sutskever. 2019.
\newblock Language models are unsupervised multitask learners.
\newblock In \emph{preprint}.

\bibitem[{Schick et~al.(2021)Schick, Udupa, and
  Schütze}]{schick2021diagnosisdebiasing}
Timo Schick, Sahana Udupa, and Hinrich Schütze. 2021.
\newblock Self-diagnosis and self-debiasing: A proposal for reducing
  corpus-based bias in nlp.
\newblock In \emph{Transactions of the Association of Computational Linguistics
  (TACL)}.

\bibitem[{Taylor et~al.(2022)Taylor, Kardas, Cucurull, Scialom, Hartshorn,
  Saravia, Poulton, Kerkez, and Stojnic}]{taylor2022galactica}
Ross Taylor, Marcin Kardas, Guillem Cucurull, Thomas Scialom, Anthony
  Hartshorn, Elvis Saravia, Andrew Poulton, Viktor Kerkez, and Robert Stojnic.
  2022.
\newblock Galactica: A large language model for science.
\newblock In \emph{preprint}.

\bibitem[{Vogel and Jurafsky(2010)}]{vogel2010learning}
Adam Vogel and Daniel Jurafsky. 2010.
\newblock \href {https://www.aclweb.org/anthology/P10-1083} {Learning to follow
  navigational directions}.
\newblock In \emph{Proceedings of the 48th Annual Meeting of the Association
  for Computational Linguistics}, pages 806--814, Uppsala, Sweden. Association
  for Computational Linguistics.

\bibitem[{Vrandečić and Krötzsch(2014)}]{vrandecic2014wikidata}
Denny Vrandečić and Markus Krötzsch. 2014.
\newblock Wikidata: a free collaborative knowledgebase.
\newblock In \emph{Communications of the ACM}.

\bibitem[{Wang et~al.(2022)Wang, Mishra, Alipoormolabashi, Kordi, Mirzaei,
  Arunkumar, Ashok, Dhanasekaran, Naik, Stap et~al.}]{wang2022super}
Yizhong Wang, Swaroop Mishra, Pegah Alipoormolabashi, Yeganeh Kordi, Amirreza
  Mirzaei, Anjana Arunkumar, Arjun Ashok, Arut~Selvan Dhanasekaran, Atharva
  Naik, David Stap, et~al. 2022.
\newblock Super-naturalinstructions: generalization via declarative
  instructions on 1600+ tasks.
\newblock EMNLP.

\bibitem[{Wei et~al.(2021)Wei, Bosma, Zhao, Guu, Yu, Lester, Du, Dai, and
  Le}]{wei2021finetuned}
Jason Wei, Maarten Bosma, Vincent Zhao, Kelvin Guu, Adams~Wei Yu, Brian Lester,
  Nan Du, Andrew~M Dai, and Quoc~V Le. 2021.
\newblock Finetuned language models are zero-shot learners.
\newblock In \emph{International Conference on Learning Representations}.

\bibitem[{Welleck et~al.(2020)Welleck, Kulikov, Roller, Dinan, Cho, and
  Weston}]{welleck2020unlikelihood}
Sean Welleck, Ilia Kulikov, Stephen Roller, Emily Dinan, Kyunghyun Cho, and
  Jason Weston. 2020.
\newblock Neural text generation with unlikelihood training.
\newblock In \emph{International Conference on Learning Representations
  (ICLR)}.

\bibitem[{Welleck et~al.(2022)Welleck, Lu, West, Brahman, Shen, Khashabi, and
  Choi}]{welleck2022selfcorrect}
Sean Welleck, Ximing Lu, Peter West, Faeze Brahman, Tianxiao Shen, Daniel
  Khashabi, and Yejin Choi. 2022.
\newblock Generating sequences by learning to self-correct.
\newblock In \emph{preprint}.

\bibitem[{Xu et~al.(2021)Xu, Pathak, Wallace, Gururangan, Sap, and
  Klein}]{xu2021minority}
Albert Xu, Eshaan Pathak, Eric Wallace, Suchin Gururangan, Maarten Sap, and Dan
  Klein. 2021.
\newblock Detoxifying language models risks marginalizing minority voices.

\bibitem[{Yang et~al.(2021)Yang, Hu, Chow, Ramadge, and
  Narasimhan}]{yang2020safe}
Tsung-Yen Yang, Michael Hu, Yinlam Chow, Peter~J. Ramadge, and Karthik
  Narasimhan. 2021.
\newblock Safe reinforcement learning with natural language constraints.
\newblock In \emph{Neural Information Processing Systems (NeurIPS)}.

\end{thebibliography}
\bibliographystyle{acl_natbib}
\newpage
\clearpage
\appendix

\section{Appendix}\label{appx:appendix}

\subsection{Data Generation Process}\label{sec:datagen}

\begin{figure*}[t]
    \centering
    \includegraphics[width=1.0\textwidth]{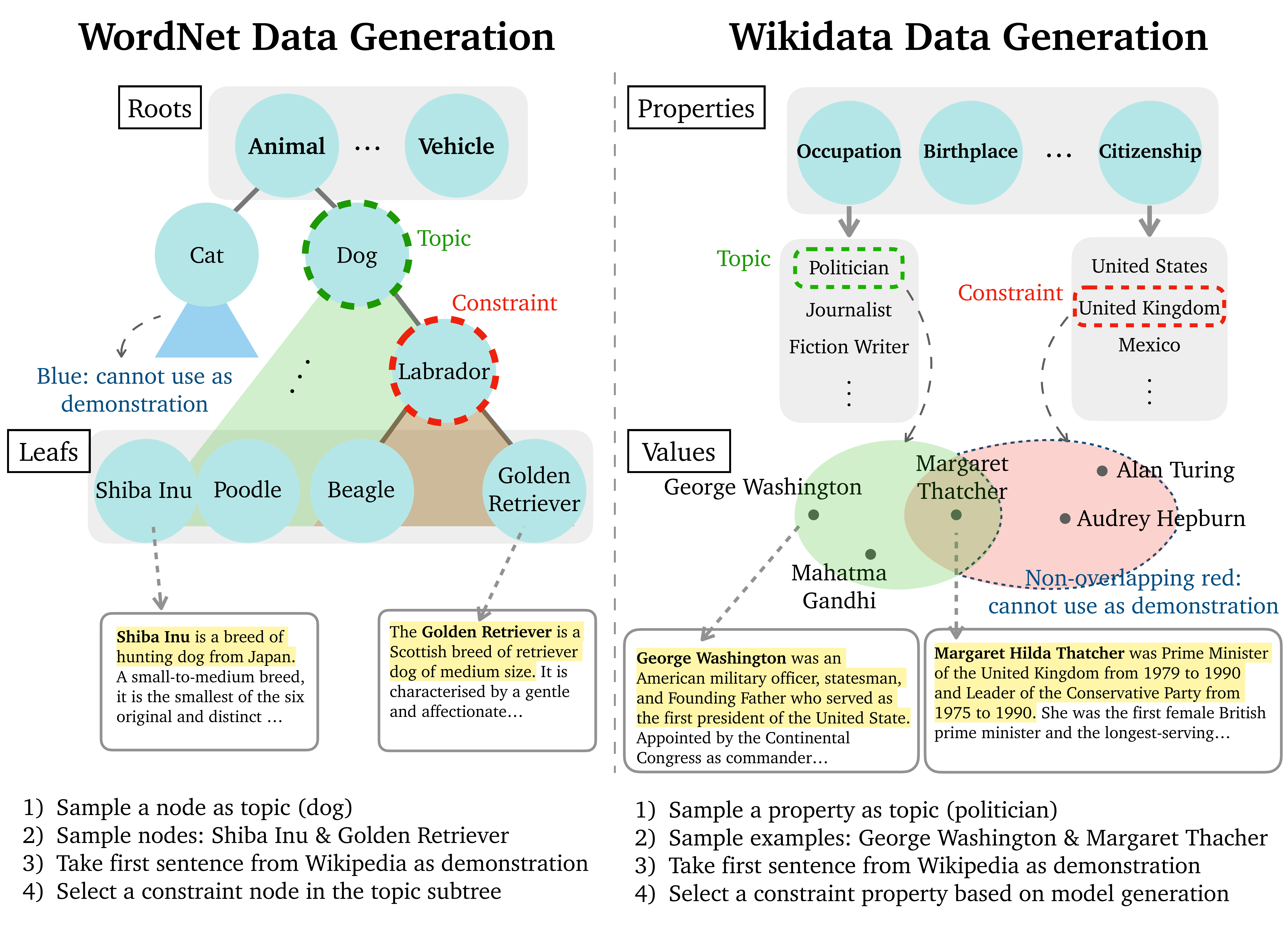}
    \caption{
    Data generation process for WordNet (left) and Wikidata (right).
    Note that in WordNet, the topic and constraint need not be connected.
    }
    \label{fig:datagen} 
\end{figure*}

Table~\ref{fig:datagen} shows how the topic and constraint are sampled from the two datasets WordNet and Wikidata.

\paragraph{WordNet.} Each example is constructed by: 1) sampling a node as the topic, 2) sampling $\lvert \xdemos \rvert = 3$ nodes under the topic node, and 3) generating a continuation from GPT2-XL \cite{radford2019gpt2} and using the generated node as the constraint. Note that both the topic and the constraint are within the same category.

\paragraph{Wikidata.} Each example is constructed by: 1) first sample a property and a value as the topic, 2) sample $\lvert \xdemos \rvert = 3$ entities from the property-value name set, and 3) generate a continuation from GPT2-XL and use the generated entity as a constraint. In contrast to WordNet, the topic and constraint are from different categories. To ensure their information does not update over time, we use only names of deceased people.

\subsection{Natural Language Instruction Templates}\label{sec:templates}
We provide the natural language instruction template examples in Table~\ref{table:nl_temp} for training (template 0-2) and development sets (template 3-5). The templates vary in their instruction positions. In template 0 and 3, the topic and constraint specification is added to the beginning and the end, respectively, with demonstration examples in the middle. On the other hand, template 1 and 4 put demonstrations to the bottom and specify the topic and constraint at once in the beginning. Note that the word use also differs between templates, sometimes

\subsection{Method Details}\label{sec:method_detail}

\makeatletter
\newcounter{phase}[algorithm]
\newlength{\phaserulewidth}
\newcommand{\setphaserulewidth}{\setlength{\phaserulewidth}}
\newcommand{\Phase}[1]{%
  \vspace{-2.60ex}
  \Statex\leavevmode\llap{\rule{\dimexpr\labelwidth+\labelsep}{\phaserulewidth}}\rule{\linewidth}{\phaserulewidth}
  \Statex\strut\refstepcounter{phase}\textit{Stage~\thephase~--~#1}%
  \vspace{-3.80ex}\Statex\leavevmode\llap{\rule{\dimexpr\labelwidth+\labelsep}{\phaserulewidth}}\rule{\linewidth}{\phaserulewidth}}
\makeatother

\algnewcommand{\LineComment}[1]{\State \(\triangleright\) #1}

\begin{figure}[t]
\centering
\begin{minipage}{\linewidth}
\begin{algorithm}[H]
\small
\caption{\sysname{} (Textual Example Guidance)}\label{algo:gen_and_guide}
\begin{algorithmic}[1]
\State Initialize $p_\text{gen}(x; \theta)$ \Comment{Pre-trained generation LM}
\State $p_\text{guide}(x; \theta) \leftarrow p_\text{gen}(x; \theta)$ \Comment{Same LM for guidance}
\\\hrulefill
\Phase{Distilling Self-Guidance}
\\\hrulefill
\State Initialize parameters $\phi$ and prefix $P$
\For {$\xin = (\xtask, \xconst, \xdemos) \in \mathcal{D}_{\text{train}}$}
    \State $y^* = p_\text{guide}(y \mid \texttt{query-template}(\topicword); \theta)$
    \State Minimize $\mathcal{L}(\phi) = - \sum_j p_\text{guide}(y_j^* \mid [\xin; P]; \theta, \phi)$
\EndFor \\
\Return $\phi(P)$ \Comment{Only saving the activations}
\\\hrulefill
\Phase{Guided Generation}
\\\hrulefill
\State Sample guidance $x^g \sim p_\text{guide}(x^g \mid \xin; \phi(P))$
\State Initialize trie $\Gamma$ using guidance $x^g$
\State Set trie tree level $l=1$
\For {$t=1 \dots T$}
    \State $\{w_k\} = \Gamma(l)$ \Comment{Retrieve a set of tokens}
    \State $s' = \texttt{Bag-of-Tokens}(\{ w_k \})$ %
    \State Obtain $s''$ for constraint following same steps
    \LineComment{Logits $s$ comes from $p_{\text{gen}}(x_j \mid x_{<j}) = \softmax(s)$}
    \State $p'_{\text{gen}}(x_j \mid x_{<j}) = \softmax(s + \alpha s' - \beta s'')$
    \State $x_t \sim p'_{\text{gen}}(x_j \mid x_{<j})$
    \If {$x_j \in \{ w_k \}$}
        \State $l = l + 1$
    \EndIf
\EndFor \\
\Return $x_1, \dots, x_T$

\end{algorithmic}
\end{algorithm}
\end{minipage}
\end{figure}

\paragraph{Training and inference details.}
During self-guidance distillation, we add for each topic and constraint $10$ prefix tokens and the MLP with hidden size $512$, and save only the activation for inference. 
We train with batch size of $16$ using the AdamW optimizer \cite{loshchilov2017decoupled} with learning rate $3e-5$ for $20$ epochs. 
During guided generation, we set $\alpha = 5.0$ and $\beta = 100.0$ and use greedy decoding. The binary verifier guidance uses $8$ tokens for greedy look-ahead. 

We provide a complete algorithm for \textguide{} in Algorithm~\ref{algo:gen_and_guide}.

\begin{table*}[th]
    \centering
    \resizebox{0.95\columnwidth}{!}{%
    \begin{tabular}{cp{0.8\textwidth}p{0.14\textwidth}}
    \toprule
    ID & Template & Instruction \newline Position\\
    \midrule
    0 & Write down examples of [topic]. & \multirow{3}{*}{Begin \& End} \\
    & [Demonstrations] & \\
    & Continue listing them but do not include examples of [constraint]. & \\
    \midrule
    1 & Below we show examples of [topic]. & \multirow{3}{*}{Begin} \\
    & Following these examples, keep listing but don’t mention [constraint]. & \\
    & [Demonstrations] & \\
    \midrule
    2 & [Demonstrations] & \multirow{3}{*}{End}\\
    & Below we show examples of [topic]. & \\
    & Following these examples, keep listing but don’t mention [constraint]. & \\
    \midrule
    3 & Generate examples that are under the category of [topic]. & \multirow{3}{*}{Begin \& End} \\
    & [Demonstrations] & \\
    & Now keep generating but exclude anything that's in the category of [constraint]. & \\
    \midrule
    4 & List out examples of [topic]. & \multirow{3}{*}{Begin} \\
    & Right after these examples, continue listing but avoid mentioning [constraint]. & \\
    & [Demonstrations] & \\
    \midrule
    5 & [Demonstrations] & \multirow{3}{*}{End}\\
    & The above are sentences describing [topic]. & \\
    & Now write similar sentences as the above while omitting any mention of [constraint]. & \\
    \bottomrule
    \end{tabular}
    }
    \caption{Natural language instruction templates in training (0-2) and development (3-5) sets.
    }
    \label{table:nl_temp}
\end{table*}

\paragraph{Top-k token guidance.}
While we only use top-k of the next token probability from the guidance distribution, we could decode multiple steps to handle multi-token entities. To encourage only generating one entity, the query template can be modified to ``What is \textit{one} example of $\constword$''. We leave this to future exploration and research.

\subsection{Qualitative Examples}
We show qualitative example of input instance and model generated output in Table~\ref{table:generation_examples}.

\begin{table*}[th]
    \centering
    \small
    \resizebox{1.0\columnwidth}{!}{%
    \begin{tabular}{cp{0.5\textwidth}p{0.5\textwidth}}
    \toprule
     Dataset & Input Instructions and Demonstrations & Generation \\
    \midrule
    WordNet & \sethlcolor{HLGreen}\hl{List out examples of bivalve.} \newline \sethlcolor{HLRed}\hl{Right after these examples, continue listing but avoid mentioning scallop.} \newline - [...] "ark shells" because species such as arca have a large flat area between the umbones [...] \newline - Placopecten magellanicus, previously listed as pecten tenuicostatus and as pecten grandis and once referred to as the "giant scallop", [...] \newline - Argopecten irradians, [...], common names atlantic bay scallop [...]  & \underline{GPT-3 Output:} \newline Aequipecten irradians, common names atlantic bay scallop or bay scallop, is a marine bivalve mollusk in the family pectinidae. \newline \newline \underline{\sysname{} Output (Ours)}: \newline Clams are a group of bivalve mollusks that are found in the northern Atlantic Ocean. \\
    \midrule
    Wikidata & \sethlcolor{HLGreen}\hl{Generate examples that are under the category of journalist.} \newline - Michael Mackintosh Foot was a British Labour Party politician who served as Labour Leader from 1980 to 1983. [...] \newline - Henry George was an American political economist and journalist. [...] \newline - John Griffith London was an American novelist, journalist and social activist. [...] \newline \sethlcolor{HLRed}\hl{Now keep generating but exclude anything that's in the category of people who were born in Cambridge.} & \underline{GPT-3 Output:} \newline Elizabeth Gurley Flynn (May 30, 1890 – September 5, 1964) was an American labor activist, anarchist and socialist/communist organizer born in Concord, New Hampshire. \newline \newline \underline{\sysname{} Output (Ours)}: \newline William F. Buckley Jr. (born William F. Buckley; June 18, 1925 – February 3, 2015) was an American conservative political commentator, author, and publisher. He was the editor of National Review [...]\\
    \bottomrule
    \end{tabular}
    }
    \caption{Example natural language instruction input and model output comparison between \sysname{} \textguide{} and GPT-3 (\texttt{davinci}). 
    }
\label{table:generation_examples}
\end{table*}

\end{document}